%% file: iclr2025_conference.tex
\documentclass{article} % For LaTeX2e
\usepackage{newtxtext} % Improved Times font support
\usepackage{iclr2025_conference,times}

\input{math_commands.tex}

\usepackage{graphicx} % Required for including images
\usepackage{caption} % For more complex captioning
\usepackage{array}
\usepackage{booktabs}
\usepackage{multirow}
\usepackage{hyperref}
\usepackage{url}
\usepackage{algorithm}
\usepackage{algpseudocode}
\usepackage{xcolor} %
\usepackage{colortbl}
\usepackage{tabularx} %
\usepackage{amssymb}
\usepackage{xcolor} %
\usepackage{colortbl}
\usepackage{tabularx} %
\usepackage{subcaption} 
\usepackage{arydshln} % 添加虚线的包
\usepackage{xcolor}   % 用于自定义颜色

\usepackage{microtype}

\usepackage{pifont}
\usepackage{tcolorbox}

\definecolor{lightblue}{HTML}{ebf3f8}

\definecolor{mediumblue}{HTML}{d7e8f2}

\definecolor{deepblue}{HTML}{c8dfed}

\usepackage{ulem}
\newcommand{\cc}[1]{{\textcolor{red}{#1}}}

\newcommand{\Dataset}{\textit{LongGenBench}}
\newcommand{\dataset}{\textit{LongGenBench}}
\newcommand{\task}{super-long-form generation}
\definecolor{kellygreen}{rgb}{0.3, 0.73, 0.09}
\definecolor{alizarin}{rgb}{0.82, 0.1, 0.26}

\definecolor{lightpurple}{RGB}{147, 112, 219} % 浅紫色
\definecolor{lightgray}{RGB}{211, 211, 211} % 浅灰色
\definecolor{lightorange}{RGB}{255, 200, 120} % 浅橘色
\definecolor{lightred}{RGB}{255, 182, 193} % 浅红色，也称为粉红色

\title{LongGenbench: Benchmarking Long-Form\\ Generation in Long Context LLMs}
\author{Yuhao Wu\textsuperscript{1}, Ming Shan Hee\textsuperscript{1}, Zhiqing Hu\textsuperscript{1} and Roy Ka-Wei Lee\textsuperscript{1} \\
\textsuperscript{1}Singapore University of Technology and Design \\
\texttt{\{wu\_yuhao,mingshan\_hee,zhiqing\_hu\}@mymail.sutd.edu.sg}\\
\texttt{roy\_lee@sutd.edu.sg}
}

\iclrfinalcopy
\begin{document}

\maketitle
\begin{figure}[h]
\centering
\label{fig:llama}
\end{figure}
\begin{abstract}

Current benchmarks like ``\textit{Needle-in-a-Haystack}'' (\textit{NIAH}), \textit{Ruler}, and \textit{Needlebench} focus on models' ability to understand long-context input sequences but fail to capture a critical dimension: the generation of high-quality long-form text. Applications such as design proposals, technical documentation, and creative writing rely on coherent, instruction-following outputs over extended sequences—a challenge that existing benchmarks do not adequately address. To fill this gap, we introduce \textit{LongGenBench}, a novel benchmark designed to rigorously evaluate large language models' (LLMs) ability to generate long text while adhering to complex instructions. Through tasks requiring specific events or constraints within generated text, \textit{LongGenBench} evaluates model performance across four distinct scenarios, three instruction types, and two generation-lengths (16K and 32K tokens). Our evaluation of ten state-of-the-art LLMs reveals that, despite strong results on \textit{Ruler}, all models struggled with long text generation on \textit{LongGenBench}, particularly as text length increased. This suggests that current LLMs are not yet equipped to meet the demands of real-world, long-form text generation. We open-source \textit{LongGenBench} to promote comprehensive evaluation and improvement in this critical area, with code and data available at \url{https://github.com/mozhu621/LongGenBench}.

\end{abstract}

\input{Threading_the_needle/intro}

%\input{Threading the needle/related_work}

\input{Threading_the_needle/our_benchmark}

\input{Threading_the_needle/expriment}

\input{Threading_the_needle/Limitation}

\input{Threading_the_needle/related_work}

\input{Threading_the_needle/conclusion}

\bibliography{iclr2025_conference}
\bibliographystyle{iclr2025_conference}
\newpage
\appendix
\input{Threading_the_needle/appendix/Symbol_Explanation_Table}

\input{Threading_the_needle/appendix/4_scenairo}

\input{Threading_the_needle/appendix/evaluton_prompt}

\input{Threading_the_needle/appendix/model}
%\input{Threading_the_needle/appendix/related_work}
\input{Threading_the_needle/appendix/Metric_example}
\input{Threading_the_needle/appendix/bad_example}

\input{Threading_the_needle/appendix/Prompt_format}

\input{Threading_the_needle/appendix/Explanation_of_Unit_Differences}

\end{document}

%% file: math_commands.tex
\usepackage{amsmath,amsfonts,bm}

\def\eqref#1{equation~\ref{#1}}
\def\1{\bm{1}}

\DeclareMathAlphabet{\mathsfit}{\encodingdefault}{\sfdefault}{m}{sl}
\SetMathAlphabet{\mathsfit}{bold}{\encodingdefault}{\sfdefault}{bx}{n}

%% file: Threading_the_needle/intro.tex
\section{Introduction}

Recent advances in large language models (LLMs) have dramatically enhanced their ability to process long text sequences, supporting applications that range from document summarization to creative writing. Leading models such as GPT-4~\citep{achiam2023gpt}, LLaMa-3.2~\citep{dubey2024llama}, and Claude 2.1~\citep{anthropic_claude_2.1} manage context windows of up to 128K tokens, with the Claude 3 series~\citep{anthropic_claude_3} handling inputs exceeding 1 million tokens. However, while much attention has been given to these models' ability to retrieve and understand long \textit{input} text sequences, far less focus has been placed on their ability to generate coherent and high-quality long-form text \textit{outputs}—a critical requirement for tasks such as design proposals and creative writing.

Long-form text generation is crucial for real-world applications that require detailed, well-structured narratives, such as document summarization \citep{kumar2024longlamp}, creative writing \citep{HuaW20, HuCLXWH22}, and comprehensive question answering \citep{StelmakhLDC22, Lee2023, bai2024longwriter}. Despite this importance, current benchmarks are limited in their ability to assess long-form generation, focusing instead on shorter text outputs ($\leq$ 2K tokens) \citep{fan-etal-2018-hierarchical, FanJPGWA19, DasigiLBCSG21}, making them unsuitable for tasks requiring outputs of $\geq$16K tokens \citep{bai2024longwriter}. The challenge is further compounded by the lack of robust methods for evaluating these long sequences. The ability to follow instructions is essential for long text generation (\textit{Reversed NIAH}\footnote{Analogous to NIAH, which involves searching for a needle (retrieval) within a long input, the reversed NIAH entails placing a specific needle (instruction-following) at a designated position within a long output.}), just as effective information retrieval is fundamental for processing long-context inputs (\textit{NIAH}\citep{needle}). However, current benchmarks do not adequately assess whether the generated text adheres to the specific directives of a prompt. For instance, a prompt may require incorporating specific information at a certain point in a lengthy document, but evaluations often fail to verify the model's compliance with such instructions. This oversight represents a significant shortcoming in benchmarking, particularly because performance under explicit constraints typically predicts outcomes in tasks with more implicit constraints, such as story generation or academic paper production. If a model struggles with explicit requirements, it is likely to underperform in scenarios with subtler constraints.

%Long-form text generation is crucial for real-world applications that require detailed, well-structured narratives, such as document summarization \citep{kumar2024longlamp}, creative writing \citep{HuaW20, HuCLXWH22}, and comprehensive question answering \citep{StelmakhLDC22, Lee2023, bai2024longwriter}. Despite this importance, current benchmarks are limited in their ability to assess long-form generation, focusing instead on shorter text outputs ($\leq$ 2K tokens) \citep{fan-etal-2018-hierarchical, FanJPGWA19, DasigiLBCSG21}, making them unsuitable for tasks requiring outputs of $\geq$16K tokens \citep{bai2024longwriter}. The challenge is further compounded by the lack of robust methods for evaluating these long sequences. In particular, existing benchmarks do not thoroughly assess whether the generated text matches the requirements of the prompt. For example, a prompt may require specific information to be inserted into the middle segment of a long-form text, but current evaluations do not check if the model successfully adheres to this requirement. Furthermore, there is no evaluation of whether the generated text remains coherent after extended generation, often resulting in contradictions between earlier and later portions of the output. These issues—adherence to prompt instructions and maintaining logical coherence over long contexts—are critical for long-form generation tasks but are absent from existing benchmarks. 

Manual evaluations, while thorough, are both costly and impractical at scale. Meanwhile, automated evaluations using "LLM-as-a-judge" methods \citep{zheng2024judging} often yield results that are difficult to interpret and may not align with human judgments, raising concerns about their reliability. This highlights the need for more specialized benchmarks capable of reliably assessing the quality of super-long-form text generation.

To address this gap, we present \Dataset, a novel benchmark designed to evaluate the quality of super-long-form text generated by long-context LLMs. Unlike existing benchmarks that primarily test retrieval or reasoning over long inputs, \Dataset~ focuses on the model’s ability to generate content that follows complex instructions over extended sequences. Our benchmark introduces tasks that reflect real-world generation challenges, such as diary writing, menu planning, and urban design, where the text must adhere to specific constraints provided in the prompt. These tasks assess whether models can correctly incorporate specific details at designated points in the text, ensuring the generated content meets the requirements laid out in the prompt. By evaluating texts up to 32K tokens, \Dataset~ is the first benchmark to systematically test the ability to generate instruction-compliant long-form content across extended lengths. Table~\ref{tab:compare_benchmark} summarizes the different benchmarks supporting long-context retrieval and generation tasks. 

The evaluation tasks are organized into four distinct scenarios: Diary Writing, Menu Design, Skyscraper Design, and Urban Planning, each with varying complexity. The scenarios involve sub-tasks such as single instance, range, and periodicity, simulating realistic constraints that a model must account for. This setup allows us to measure the model’s ability to generate detailed, contextually rich outputs that satisfy a wide array of criteria.

%Our results indicate that while existing long-context LLMs excel at processing lengthy inputs, they exhibit significant performance drops when tasked with generating super-long-form outputs (see Figure \ref{fig:instruct1}). None of the models tested on \Dataset~ were able to maintain coherence or adhere to the task instructions at 16K or 32K token lengths, revealing the current limitations of long-form text generation in large-scale LLMs. We conclude by highlighting future directions for research and the importance of further developing models that can manage not only long input sequences but also coherent long output sequences. 

%\begin{figure}[t]
%	\captionsetup{type=figure} % Ensures the caption is treated as a figure caption
%	\centering
%	\includegraphics[width=0.73\linewidth, height=3.5cm]{Threading_the_needle/fig/instruct_2.pdf} % Updated path to be more generic
	%\vspace{-8pt}
%	\caption{ % Cleaned up the caption for clarity and formatting
%		\label{fig:instruct1}\footnotesize
%		The left side of the figure displays the average performance results of the model on Ruler \citep{hsieh2024ruler}, whereas the right side shows the model's average performance on the \dataset~ (16K).
%	}
%\end{figure}

In summary, our major contributions are as follows:
\begin{itemize}
    \item To the best of our knowledge, this is the first study to address the challenge of super-long-form generation in long-context language models, highlighting the critical importance of generating coherent, high-quality text in extended contexts. 
    \item We introduce \Dataset~, a comprehensive dataset that provides a diverse set of tasks specifically designed to evaluate the super-long-form generation capabilities of LLMs across varying token lengths (16K and 32K) and levels of text complexity.
    \item We perform extensive experiments on both open-source and closed-source models, revealing that despite their advanced capabilities, most models struggle significantly with super-long-form generation tasks, particularly in maintaining instruction adherence and coherence over long outputs.
\end{itemize}

% (i) To the best of our knowledge, this is the first study to address the challenge of super-long-form generation in long-context language models, highlighting the critical importance of generating coherent, high-quality text in extended contexts. (ii) We introduce \Dataset~, a comprehensive dataset that provides a diverse set of tasks specifically designed to evaluate the super-long-form generation capabilities of LLMs across varying token lengths (16K and 32K) and levels of text complexity. (iii) We perform extensive experiments on both open-source and closed-source models, revealing that despite their advanced capabilities, most models struggle significantly with super-long-form generation tasks, particularly in maintaining instruction adherence and coherence over long outputs.

\input{Threading_the_needle/tabel/Compare_benchmark}
%In summary, our major contributions are as follows:
%\begin{itemize}
%    \item To the best of our knowledge, this is the first study to introduce the challenge of \task~ in long-context language models, emphasizing the importance of generation within these extended contexts.
%    \item We introduce \Dataset~, a dataset that includes tasks for evaluating the \task~ capabilities of LLMs across multiple token lengths (16K and 32K) and various text depth ranges.
%    \item We conducted comprehensive experiments on both open-source and closed-source models. Our findings indicate that most models fail to effectively perform the \task~ tasks, despite their claimed capabilities. 
%\end{itemize}

%% file: Threading_the_needle/tabel/Compare_benchmark.tex
\begin{table}[t]
\centering
\small
\caption{Comparison of Long-Context LLM Benchmarks. For the retrieval tasks' datasets, we measure length based on the number of processing tokens, while for the generation tasks' datasets, we calculate the average number of generation words produced by LLMs. `Long-length' indicates if LLMs to analyze or generate text that is at least 8K token.}
\resizebox{0.85\textwidth}{!}{
\begin{tabular}{@{}clrcc@{}}

\toprule
Type of Task & Benchmark & Type of Data & Avg Len & Long-Length  \\

\midrule
\multirow{3}{*}{Retrieval} 
 & Longbench\citep{longbench} & hybrid & $ \sim $8k & \textcolor{green}{\ding{51}}  \\
& NIAH\citep{needle} &  synthetic & Any & \textcolor{green}{\ding{51}}   \\
 & Ruler\citep{hsieh2024ruler} &  synthetic & Any & \textcolor{green}{\ding{51}}    \\
\midrule
\multirow{3}{*}{Generation} & ELi5\citep{fan-etal-2019-eli5} & hybrid & $ \sim $0.2K  & \textcolor{red}{\ding{55}} \\
 %& Hellobench\citep{Que2024HelloBenchEL} & hybrid & $ \sim $2K & - & \textcolor{red}{\ding{55}} \\
 & Longwrite\citep{bai2024longwriter} & synthetic & $ \sim $2.7K & \textcolor{red}{\ding{55}}  \\
 & \textbf{\Dataset(Ours)} &  synthetic & $ \sim $20K & \textcolor{green}{\ding{51}}  \\
\bottomrule
\end{tabular}}
\label{tab:compare_benchmark}
\end{table}

%% file: Threading_the_needle/our_benchmark.tex
\section{\Dataset~Benchmark}
\label{dataset}

%In this section, we provide a detailed examination of our benchmark as outlined, encompassing Task definition, Different Scenario setups, Specific task instruction, Evaluation metric and Evaluations pipeline.
\begin{figure}[t]
	\captionsetup{type=figure} % Ensures the caption is treated as a figure caption
	\centering
	\includegraphics[width=0.85\linewidth]{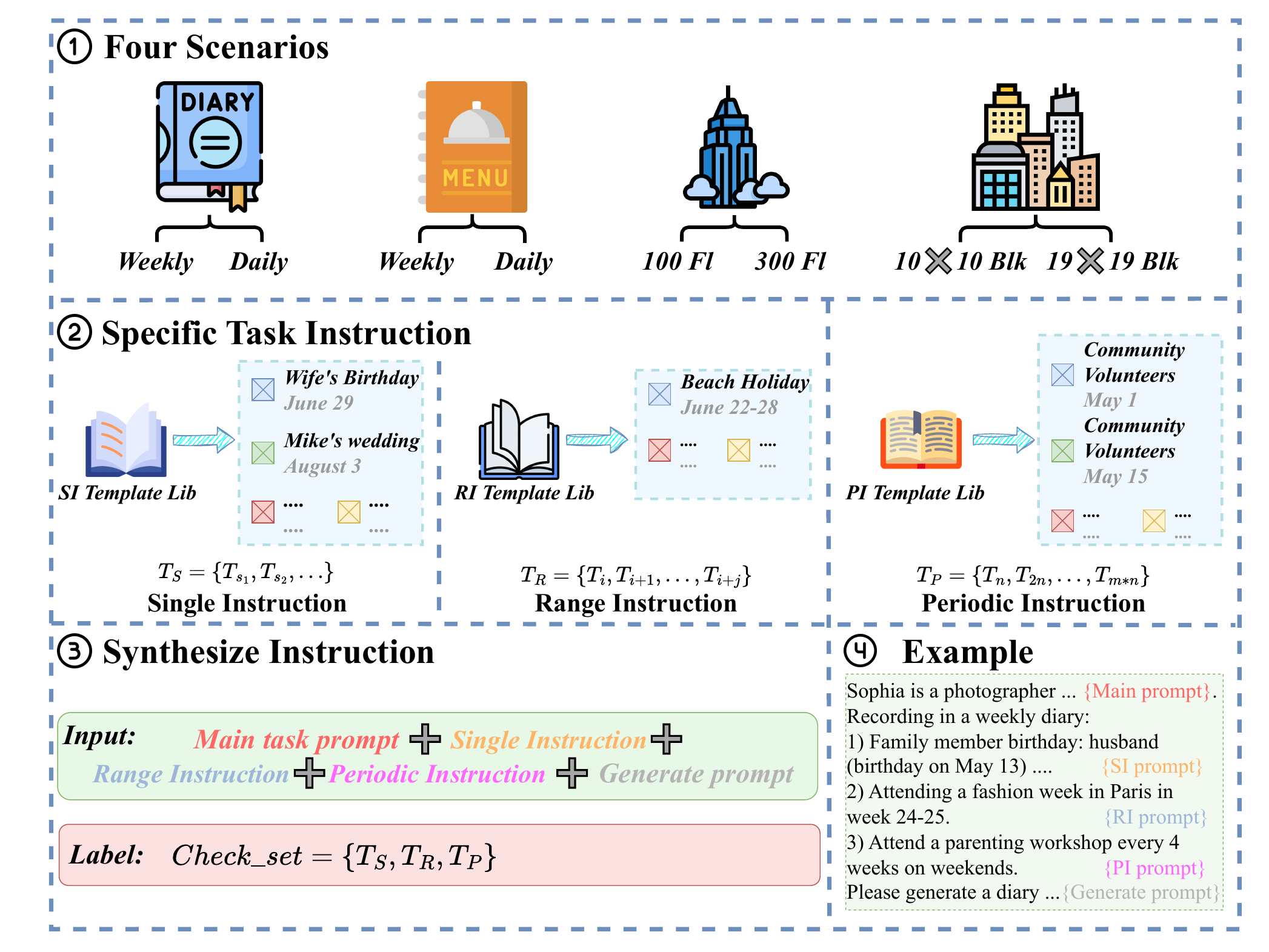} % Updated path to be more generic
	\vspace{-8pt}
	\caption{ \small{
 \textbf{\dataset~Overview:}
1) \textbf{Scenario Selection:} Select from four scenarios—Diary, Menu Design, Skyscraper Design, and Urban Planning—each offered in both short and long versions to determine the main task prompt.
2) \textbf{Task Instruction:} Employ the template libraries SI (Single), RI (Range), and PI (Periodic) to generate tasks characterized by random times or locations, along with the corresponding prompts and verification sets.
3) \textbf{Instruction Synthesis:} Integrate all prompts generated in the prior step to create a comprehensive set of instructions with a final check-set.
4) \textbf{Example:} An illustration of Sophia's weekly diary task is provided as an example.}
		}
        \label{fig:benchmark}

\end{figure}

\subsection{Task Definition} 
Evaluating the quality of \task~ presents a unique set of challenges due to the inherent complexity of long texts. Traditional human evaluation methods, while precise, are expensive and not scalable. Although using large language models for assessment is feasible, their lack of interpretability often hampers their utility. Thus, we focus on the "instruction-following" task in \task, where the most must include specific details in the generated text. This task reflects real-world scenarios that require a high degree of attention to detail over extended sequences, such as technical documentation or detailed design proposals. In this study, we define a task type termed \textit{Strictly Sequential Tasks}, which involves the sequential completion of subtasks $\mathbf{T} = \{T_1, T_2, T_3, \ldots, T_n\}$\footnote{In Appendix \ref{app:Symbol Definitions and Descriptions}, there is a detailed description of the definitions of mathematical symbols.}, where each subtask is responsible for generating a specific volume of text. For instance, an LLM might be tasked with designing a 100-floor skyscraper, specifying the content and purpose of each floor.

%Evaluating the quality of \task~ is both crucial and challenging, given the inherent complexity of long texts. Traditional human evaluation methods, while precise, are expensive and not scalable. Although feasible, using large language models for assessment often lacks interpretability. Thus, we focus on the ``\textit{instruction following}"''task in \task. In this study, we define a task type termed \textit{Strictly Sequential Tasks}, which involves the sequential completion of subtasks $\mathbf{T} = \{T_1, T_2, T_3, \ldots, T_n\}$, where each subtask is responsible for generating a specific volume of text. For instance, an LLM might be tasked with designing a 100-floor skyscraper, specifying the content and purpose of each floor.

%\subsection{Task Definition} Evaluating the quality of \task~ is both crucial and challenging, given the inherent complexity of long texts. Traditional human evaluation methods, while precise, are expensive and not scalable. Although feasible, using large language models for assessment often lacks interpretability. Thus, we focus on the "instruction following" task in \task~. 
%In this study, We define a task type termed \textit{Strictly Sequential Tasks}, which involves the sequential completion of subtasks $\mathbf{T} = \{T_1, T_2, T_3, \ldots, T_n\}$, where each subtask is responsible for generating a specific volume of text. For instance, an LLM might be tasked with designing a 100-floor skyscraper, specifying the content and purpose of each floor.

\subsection{Four Distinct Scenario Setups}
To comprehensively assess the long-form generation capabilities of models, we have devised four distinct task scenarios to supplement our predefined tasks, as illustrated in Figure \ref{fig:benchmark} (1). These scenarios fall into two categories: Temporal (Diary Writing, Menu Design) and Spatial (Skyscraper Design, Urban Planning). Moreover, each scenario incorporates both short and long versions to assess the effectiveness of various output lengths.

These scenarios were carefully chosen to reflect both creative and technical long-form generation tasks. They offer a diverse set of challenges by including temporal tasks (e.g., Diary Writing) that require maintaining consistent information over time and spatial tasks (e.g., Urban Planning) that test the model’s ability to handle spatial relationships and detailed designs. These scenarios mirror real-world applications, from planning documents to creative writing, and thus provide a comprehensive evaluation of long-context LLMs. Table \ref{tab:Four_scenarios} offers comprehensive descriptions for each scenario, with each designed around a unique template to generate 100 different task instructions\footnote{Examples of Task Instructions for each scenario are provided in Appendix \ref{SCENARIO}.}.

\input{Threading_the_needle/tabel/Four_scenairo}

%\subsection{Four Distinct Scenario Setups}
%In conjunction with the design of a 100-floor skyscraper, we have devised three distinct task scenarios to supplement our predefined tasks, as illustrated in Figure \ref{fig:benchmark} (1). These scenarios fall into two categories: Temporal (Diary Writing, Menu Design) and Spatial (Skyscraper Design, Urban Planning). Moreover, each scenario incorporates both short and long version to assess the effectiveness of various output lengths. Each scenario is carefully engineered to evaluate the model’s capacity to produce context-sensitive and detailed lengthy texts tailored to our specific task requirements. Table \ref{tab:Four_scenarios} offers comprehensive descriptions for each scenario, with each designed around a unique template to generate 100 different task instructions\footnote{Examples of Task Instructions for each scenario are provided in Appendix \ref{SCENARIO}.}.
%\input{Threading_the_needle/tabel/Four_scenairo}

\subsection{Specific Task Instruction} 
To enhance task control and flexibility, we have developed three distinct task settings:
\begin{itemize}
    \item \textbf{Single Instruction} (\textit{SI}): Injects specific information at a unique point within the generated text.
    \[ T_S = \{T_{s_1}, T_{s_2}, \ldots\}\]
    \item \textbf{Range Instruction} (\textit{RI}): Requires the model to incorporate information within specified ranges of the text. 
    \[ T_R = \{T_{R_i}, T_{R_{i+1}}, \ldots, T_{R_{i+j}}\} \]
    \item \textbf{Periodic Instruction} (\textit{PI}): Distributes information at predefined intervals throughout the text.
    \[ T_P = \{T_{P_n}, T_{P_{2n}}, \ldots, T_{P_{m*n}}\}\]
    \item \textbf{Check Set}: Includes tasks for all three aforementioned settings. 
    \[ Check\_set = \{T_S, T_R, T_P\} \]
\end{itemize}

For example, in the design of a 100-floor skyscraper, the \textit{Single Instruction} may specify that the 34th floor hosts an aerial gym and the 54th floor houses a law firm. The \textit{Range Instruction} might designate floors 1 through 9 as a comprehensive shopping mall, whereas the \textit{Periodic Instruction} could dictate that starting from the 20th floor, every 10th floor incorporates a small aerial garden. 

We utilize over 20 templates for each type of instruction, with the floors or locations being randomly assigned to ensure task diversity. These settings, applied via various templates, guarantee controlled coverage across all textual positions, thus facilitating a comprehensive and efficient evaluation, as illustrated in Figure \ref{fig:benchmark} (2).

Through this approach, we generate the main task instructions $T$ and simultaneously acquire the corresponding $Check\_set$, which supports subsequent evaluations and constructs a task conducive to \task. Subsequently, we splice the main task prompt with the specific task instructions (STI)\footnote{Each of our main task instructions $T$ splice 5 single instructions, 1 range instruction task, and 1 periodic instruction task.} and add the generation prompt to form the final evaluation data.

\subsection{evaluation Metric}
\label{Sec:Metric}

To quantitatively evaluate performance for \dataset~tasks, we introduce three complementary metrics:

\paragraph{Main Task Completion.}
This metric evaluates the extent to which all designated subtasks are accomplished. The completion rate is quantified using the following equation:
\[
\text{Completion Rate (CR)} = \frac{\text{Number of Completed Subtasks}}{\text{Total Number of Subtasks}} \times 100\%
\]
In this context, the numerator denotes the count of subtasks successfully executed by the model, and the denominator represents the total number of subtasks defined in the Strictly Sequential Task. For instance, does the model consistently complete a diary entry for each day without omitting any dates?

\paragraph{Specific Task Instruction Completion (STIC-1).}
This metric evaluates the model’s adherence to specific task instructions. We calculate the completion counts for the \textit{Single Instruction} (SI), \textit{Range Instruction} (RI), and \textit{Periodic Instruction} (PI). STIC-1 quantifies how well the model follows these instructions across subtasks, focusing on whether the instructions are correctly implemented. For example, in the Skyscraper Design task, if the model is instructed to place an aerial gym on the 34th floor and consistently places it on a different floor, it would receive a lower STIC-1 score.

\[
{
  \text{STIC-1} =  \frac{\text{Single Instruction} + \text{Range Instruction} + \text{Periodic Instruction}}{\text{Total Number of \textbf{Outputs} to Specific Task Instructions}} 
}
\]

\paragraph{Specific Task Instruction Completion-2 (STIC-2).}
STIC-2 provides a more granular assessment by measuring the overall completion of specific task instructions, including their presence and execution quality across all subtasks. In addition to adherence, it assesses whether the model consistently follows these instructions throughout the entire task. For instance, if the model periodically repeats certain elements but not at the required intervals, it would affect its STIC-2 score.

\[
{
  \text{STIC-2} =  \frac{\text{Single Instruction} + \text{Range Instruction} + \text{Periodic Instruction}}{\text{Total Number of Specific Task Instructions}} 
}
\]

STIC-1 is primarily concerned with the completion rate of instructions that result in sub-scenarios, focusing on whether instructions are correctly executed. In contrast, STIC-2 assesses the overall completion of the specific instruction task, including the presence of sub-scenarios and their completion status\footnote{In Appendix \ref{app:STIC-1 & STIC-2}, we provide a detailed explanation of STIC-1 and STIC-2, along with a case study analysis.}.

\subsection{evaluations Pipeline}
Our evaluation process follows a structured pipeline: First, we use a long-context LLM to complete the task instruction \(T\), generating an answer \(A\), which is then divided into sub-tasks as \(A = \{A_1, A_2, \ldots, A_n\}\). Next, based on the specific instructions in the \textit{check\_set}, we identify the relevant sub-tasks within \(A\). Finally, we evaluate each sub-task by \(\text{eval}(A_i, T_i)\) to compute the final completion score, as detailed in Algorithm \ref{alg: pipeline}. This pipeline ensures that the evaluation is both systematic and comprehensive, assessing the model’s performance across different instruction settings and levels of complexity \footnote{In Appendix \ref{app: evaluation prompts}, there is a detailed evaluations pipeline and example.}. While \textit{LongGenBench} primarily evaluates the model's ability to follow detailed instructions, future work could expand the benchmark to include more open-ended tasks that assess creativity and logical reasoning. This would provide a broader evaluation of a model's capabilities in generating coherent, engaging, and logically sound long-form text.

% Our evaluation process consists of the following steps: 1) First, we use a long-context language model (LLM) to complete the aforementioned task instruction \(T\) and obtain its Answer \(A\), which is then divided into sub-tasks as \(A = \{A_1, A_2, \ldots, A_n\}\). 2) Next, based on the instructions that need to be evaluated in the \textit{check\_set}, we identify the corresponding sub-tasks in the Answer. 3) Finally, we evaluate each sub-task by \(\text{eval}(A_i, T_i)\) to calculate the final completion score, as specified by the algorithm \ref{alg: pipeline}.

\begin{algorithm}
\caption{Evaluations Pipeline}
\label{alg: pipeline}
\begin{algorithmic}[1]
\small
\Statex $\textbf{Initialization:}$
\State $\text{Task instructions} \rightarrow T$
\State $\text{Tested long context LM} \rightarrow \text{model}$
\State $\text{Set of Special Task Instruction for evaluation matching} \rightarrow \text{Check\_Set}$
\Statex

\Statex $\textbf{Main Process:}$
\State $\text{Use Tested model to get Answer for } T \rightarrow A$
\State $A \rightarrow \{A_1, A_2, \ldots, A_m\} \text{, split into subtasks}$

\State $\text{empty set for storing evaluations} \rightarrow E$

\For{each $T_i$ in \text{$Check\_Set$}}
    \If{there is $A_i \text{ matching } T_i$}
        \State $\text{eval}(A_i, T_i) \rightarrow E_i$
        \State $E \rightarrow \text{Add } E_i \text{ to } E$
    \EndIf
\EndFor

\State $\sum E \rightarrow Score \text{, compute the final completion score}$
\State \Return $Score$
\end{algorithmic}
\end{algorithm}

% \paragraph{Evaluation Methodology}
% To facilitate the evaluation, we trained a small BERT model for this simple classification task and also conducted a few-shot comparison using llama-3-8b-chat. The results of these comparisons are presented in Appendix X. The small BERT model has proven to be sufficiently effective for these basic tasks, eliminating the need for larger models. Thus, with the results from all the large language models (LLMs) on a test dataset of 700 samples, evaluation can be rapidly performed on a single A100 GPU in XXX minutes.

%\subsection{visual interpretation}

%% file: Threading_the_needle/tabel/Four_scenairo.tex
\begin{table}[ht]
\centering
\small
\caption{Scenario task descriptions}
\resizebox{0.8\textwidth}{!}{
\begin{tabular}{@{}cccc@{}}

\toprule
\textbf{Category} & \textbf{Scenarios} & \textbf{Task} & \textbf{Task Description} \\
\midrule
\multirow{4}{*}{Temporal} & \multirow{2}{*}{Diary} & Weekly Diary & Generate entries for each week of the year \\
 &  & Daily Diary & Generate entries for each day of the year \\
\cmidrule(l){2-4}
 & \multirow{2}{*}{Menu} & Weekly Menu & Plan menus for each week of the year \\
 &  & Daily Menu & Plan menus for each day of the year \\
\midrule
\multirow{4}{*}{Spatial} & \multirow{2}{*}{Skyscraper Design} & 100-floor Design & Develop a design for a 100-floor skyscraper \\
 &  & 361-floor Design & Develop a design for a 300-floor skyscraper \\
\cmidrule(l){2-4}
 & \multirow{2}{*}{Urban Planning} & 10x10 block Design & Design an urban layout on a 10x10 block grid \\
 &  & 19x19 block Design & Design an urban layout on a 19x19 block grid \\
\bottomrule
\end{tabular}}
\label{tab:Four_scenarios}
\end{table}

%% file: Threading_the_needle/expriment.tex
\section{Experiments}

%\subsection{Experimental Setup} 
%\paragraph{Models \& Inference setup}We selected ten long-context large language models (LLMs), comprising eight open-source and two closed-source models. These models range in size from 7B to 72B parameters and feature a Mixture of Experts (MoE) architecture. The claimed context lengths of these models vary from 32K to 128K tokens\footnote{Detailed specifications of these models are provided in Appendix \ref{models}.}. We evaluated all models using the vLLM~\citep{vllm} serving system, which incorporates efficient key-value (KV) cache memory management. Inferences were performed using BFloat16 precision on $8 \times$ NVIDIA A800 GPUs, employing greedy decoding.

\subsection{Experimental Setup} 
\paragraph{Models.} We selected ten long-context large language models (LLMs), comprising eight open-source and two closed-source models. These models range in size from 7B to 72B parameters, with one featuring a Mixture of Experts (MoE) architecture. The claimed context lengths of these models vary from 32K to 128K tokens\footnote{Detailed specifications of these models are provided in Appendix \ref{models}.}. These models were selected to represent a diverse array of architectures, covering both Mixture of Experts and standard transformer designs, as well as a range of parameter sizes. This diversity ensures a comprehensive evaluation of their ability to handle long-context tasks.

\paragraph{Inference Setup.} We utilized the vLLM~\citep{vllm} system, which optimizes key-value (KV) cache memory for efficient large-scale inference. This system is crucial for handling long-form generation efficiently, reducing memory overhead, and maximizing inference throughput. Inferences were performed using BFloat16 precision on $8 \times$ NVIDIA A800 GPUs, employing greedy decoding to generate the outputs. This setup ensured consistency and efficiency in the inference process.

%\paragraph{Task Configurations} For each scenario, we generated 800 examples at specified lengths (16K, 32K) using the designated chat template for each model. To ensure relevance and prevent off-topic responses or refusals to answer, we prefixed each task input with a specific answer prompt. We assessed model performance using the three previously defined metrics.

\paragraph{Task Configurations.} For each scenario, we generated 800 examples at two specified lengths: 16K tokens and 32K tokens. The generation was based on designated templates for each model, ensuring task-specific relevance. The tasks were selected to reflect both creative and technical long-form generation challenges, such as diary writing, urban planning, and skyscraper design. To ensure the relevance of the generated content and prevent off-topic responses or refusals to answer, we prefixed each task input with a carefully curated answer prompt designed to guide the model’s output. The tasks were specifically selected to test the models' ability to generate instruction-following long-form content in both creative and technical contexts. For example: In the \textit{Urban Planning} task, models were tasked with generating a detailed plan for a new urban district, including descriptions of key facilities such as parks, schools, and transportation systems.
% \begin{itemize}
%     \item In the \textbf{Diary Writing} task, models were required to simulate daily personal entries over a year, testing their ability to maintain consistency over long temporal spans.
%     \item In the \textbf{Urban Planning} task, models were tasked with generating a detailed plan for a new urban district, including descriptions of key facilities such as parks, schools, and transportation systems.
%     \item In the \textbf{Skyscraper Design} task, models were instructed to generate descriptions of each floor in a 100-floor building, testing their ability to manage spatial relationships and adhere to specific design constraints.
% \end{itemize}

% These tasks were chosen to represent a variety of real-world applications that require detailed, coherent long-form text generation. However, additional tasks could be considered in future studies to cover an even broader spectrum of applications, such as legal document drafting or research report generation. 

\paragraph{Evaluation Metric.} We evaluated model performance using the three metrics defined in Section \ref{Sec:Metric}: \textit{Main Task Completion}, \textit{Specific Task Instruction Completion-1 (STIC-1)}, and \textit{Specific Task Instruction Completion-2 (STIC-2)}. These metrics provided a comprehensive assessment of the models' ability to adhere to instructions and generate coherent long-form text.

\subsection{Main Result}
\input{Threading_the_needle/tabel/main_result}

The results of the long-form text generation tasks for both Short-version (16K)  and Long-version (32K) tokens are summarized in Table \ref{tab:comparison_models}.
%The models exhibited significant performance variations, highlighting differences in architecture, parameter size, and training data.

\textbf{Main Task Completion.} Significant disparities in performance across models primarily stem from differences in architecture and training datasets. Notably, models with varying parameter sizes, such as Llama3.1-8B-instruction~\citep{dubey2024llama} (under 10 billion parameters), Qwen-72B~\citep{yang2024qwen2} (over 20 billion parameters), and GPT-4o-mini~\citep{gpt-4o-mini} (a closed-source model), have demonstrated superior efficacy, successfully completing most primary tasks in full. In contrast, some models struggle with these tasks, exhibiting limitations such as: 1) models responding solely to specified subtasks, neglecting others, and 2) models halting after only completing the initial task segment, despite prompts requiring full sequential subtask completion. This issue may originate from the current instructional tuning data, which could cause partial responses in complex, lengthy tasks. Especially in GPT-4o\citep{gpt4}, it recognizes that this task will generate a long output and only provides a few examples.

\textbf{STIC-1 and STIC-2.} The \texttt{STIC-1} metric revealed strong performance in adhering to task instructions for models like \texttt{Mixtral-8x7B} and \texttt{GPT-4o-mini}, particularly in shorter sequences. However, a significant drop in \texttt{STIC-2} scores for several models indicates that maintaining instruction adherence over longer text sequences remains a challenge. This performance degradation emphasizes the need for better tuning and architectural modifications to improve long-term coherence. The MoE model, Mistral-8x7B generally outperformed dense counterparts in instruction-following over extended sequences, but both model types struggled with STIC-2 in longer generations.

A common failure mode observed across multiple models was the tendency to forget or misinterpret instructions as the sequence length increased. For example, in the \textit{Skyscraper Design} task, some models correctly described the initial few floors but deviated from the original plan as the task progressed, particularly in the 32K token setting. This highlights the memory retention issue in long-context models, which often leads to a loss of coherence and adherence to task instructions. Examples of failures where models struggled to follow instructions are provided in Appendix ~\ref{app:bad_example}.

\paragraph{Length (Number of words).}
We calculated the average output word count for models that consistently completed all subtasks, achieving at least an 80\% completion rate in sub-scenarios, excluding data from unsuccessful attempts. Most models substantially exceeded previous benchmarks for long-form generation tasks in terms of output length. Notably, the \texttt{LongWriter}~\citep{bai2024longwriter} model excelled, efficiently meeting word count requirements for each subtask. Given the results and the weighted average score (wAvg) at a sequence length of 16K, the open-source \texttt{Qwen2-72B} and the closed-source \texttt{GPT-4o} models demonstrated optimal performance. At a sequence length of 32K, the \texttt{Llama3.1-8B} model, outperformed models with larger parameters, highlighting its efficiency in managing extended lengths.
%\footnote{\cc{The unit for claim length is tokens and the unit for length is words. The reasons are detailed in Appendix \ref{app:Explanation of Unit}}.}.

%\subsection{Accuracy Trend with Varying Sequence Length}
%As show in Fig. \ref{fig:Accuracy Trend}, there is a noticeable decline in model performance over time. As the outputs become longer, all models gradually fail to adhere to the initial instructions, resulting in a performance degradation analogous to trends observed in the NIAH dataset. This deviation is particularly pronounced when the output exceeds 4,000 tokens, where adherence to instructions significantly diminishes, with marked deterioration at the 4,000-token threshold and substantial ineffectiveness by 16,000 tokens. Conversely, in tasks such as those involving NIAH and more complex multi-needle tasks of similar lengths, these models demonstrate near-perfect performance. Such disparities in performance cannot be easily explained by prevailing theories, such as "loss in the middle"~\citep{lostmiddle}, especially since the instructions are positioned within the first 500 tokens.

\subsection{Accuracy Trend with Varying Sequence Length}
\begin{figure*}[t]
	\captionsetup{type=figure}
	\centering
	\includegraphics[width=1\linewidth]{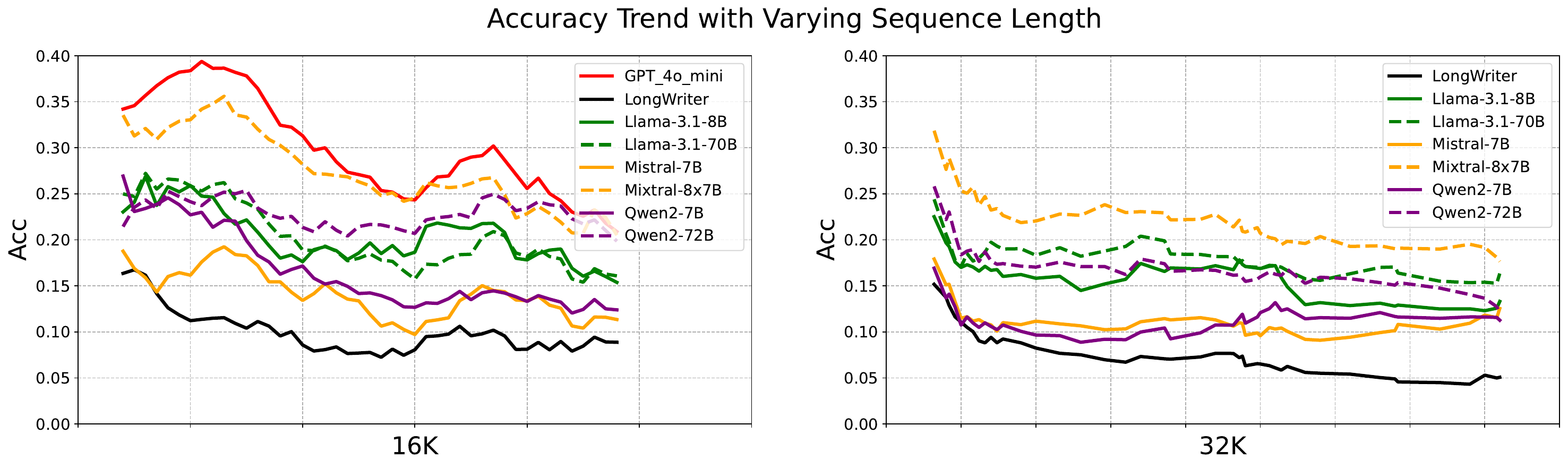}
	\caption{
		\label{fig:Accuracy Trend} 
The right side of the figure illustrates the model's performance on specific instruction tasks at 16K as sequence length increases, whereas the left side depicts performance at 32K. All curves have been smoothed with a Moving Average.
	}
\end{figure*}

As illustrated in Figure \ref{fig:Accuracy Trend}, there is a clear decline in model performance as output length increases. Models exhibited strong adherence to initial instructions at shorter sequence lengths, but performance gradually degraded as the text generation extended beyond the 4,000-token threshold. This degradation aligns with trends identified in the NIAH dataset and underscores the challenge of maintaining instruction adherence and coherence over long outputs.

This deviation becomes particularly pronounced when outputs exceed 4,000 tokens, where adherence to instructions significantly diminishes, and further deterioration is observed as outputs approach 16,000 tokens. In contrast, tasks involving shorter outputs, such as those in the NIAH dataset or simpler multi-needle tasks, showed near-perfect performance, highlighting the disparity in model behavior across different sequence lengths.

Potential reasons for this decline include limitations in the self-attention mechanism used in transformers, which may struggle to maintain meaningful context over long sequences. Additionally, models trained with limited long-form data may overfit to shorter patterns, leading to a loss of coherence in extended generations. These findings suggest that architectural changes or improved training strategies may be necessary to overcome these challenges in future iterations of LLMs.

%\subsection{Three Specific Task Instruction}

%In Figure \ref{fig:instruct_2_left}, we present the performance metrics across various tasks. The model shows similar, yet not identical, proficiency in single and range tasks, suggesting its effective execution of direct instructions for these task types. This reduction in performance can be ascribed to the additional cognitive load associated with interpreting periodic instructions, such as identifying relevant weeks in statements like "every four weeks starting from week 10." Consequently, outcomes for periodic tasks are substantially poorer compared to those for single and range tasks, which have well-defined parameters. Typically, the model's performance hierarchy is represented as $single > range > periodic$.

\subsection{Three Specific Task Instructions}

Figure \ref{fig:instruct_2_left} presents the model's performance metrics across various task types: single, range, and periodic. The model demonstrated comparable proficiency in both single and range tasks, reflecting its capability to follow direct and straightforward instructions effectively. However, the slight reduction in performance for range tasks suggests that additional complexity, such as processing multiple data points within a defined range, introduces a marginal increase in cognitive load for the model.

The most significant decline in performance was observed in periodic tasks, where the model struggled to interpret instructions that required recurring events, such as "every four weeks starting from week 10." These tasks demand a higher degree of reasoning and temporal awareness, which may challenge the model’s capacity to maintain consistency over extended sequences. As a result, outcomes for periodic tasks were considerably poorer compared to single and range tasks, which have clearer and more well-defined parameters. The model's performance hierarchy can generally be summarized as $single > range > periodic$, highlighting the increased difficulty associated with periodic tasks. This trend underscores the need for future improvements in long-context models, particularly in handling more complex, time-based instructions.

%\subsection{Compare with long context input }
%In this subsection, we examine the relationship between a model's ability to handle long-range inputs and its performance on long-range outputs. Specifically, we investigate whether a model's capacity to manage long-range inputs corresponds to improved performance on long-range outputs. For this analysis, we use the RULER dataset.
%RULER is a novel synthetic benchmark designed to flexibly configure sequence length and task complexity, making it ideal for comprehensive evaluations of long-context language models. We compare the models' performance on sequences of the same length, as shown in Figure \ref{fig:instruct_2_right}, which indicates a certain degree of correlation between input handling and output performance. At 16K, the Pearson correlation coefficient is 0.51, while at 32K, it increases to 0.66, highlighting a significant overlap in their capabilities.

\subsection{Comparison with Long-Context Input}
\begin{figure*}[t]
	\captionsetup{type=figure}
	\centering
	\begin{minipage}[t]{0.45\linewidth}
		\centering
		\includegraphics[width=\linewidth]{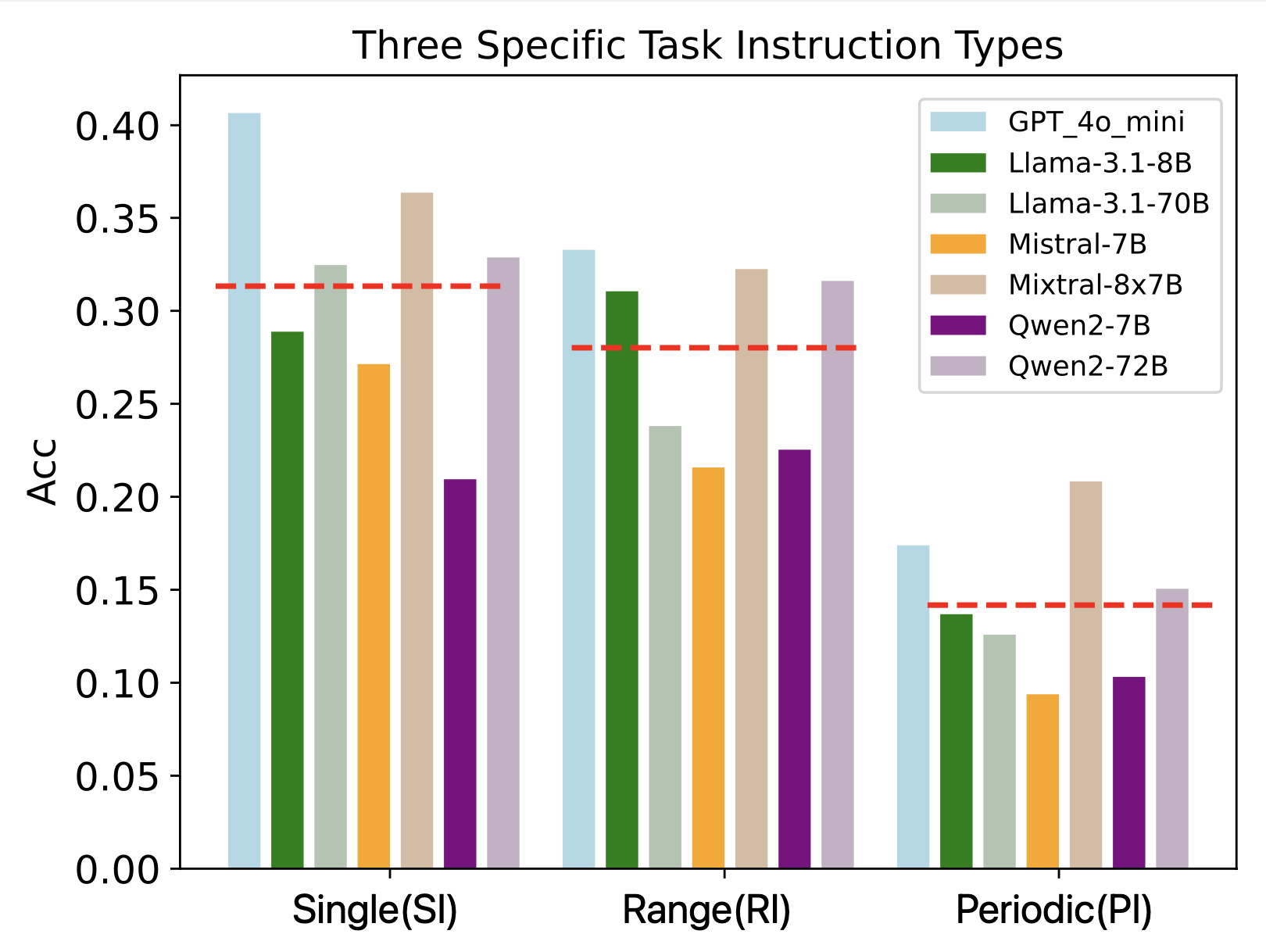}
		\subcaption{\footnotesize{Performance Comparison on three tasks settings}}
		\label{fig:instruct_2_left}
	\end{minipage}
	\hfill
	\begin{minipage}[t]{0.45\linewidth}
		\centering
		\includegraphics[width=\linewidth]{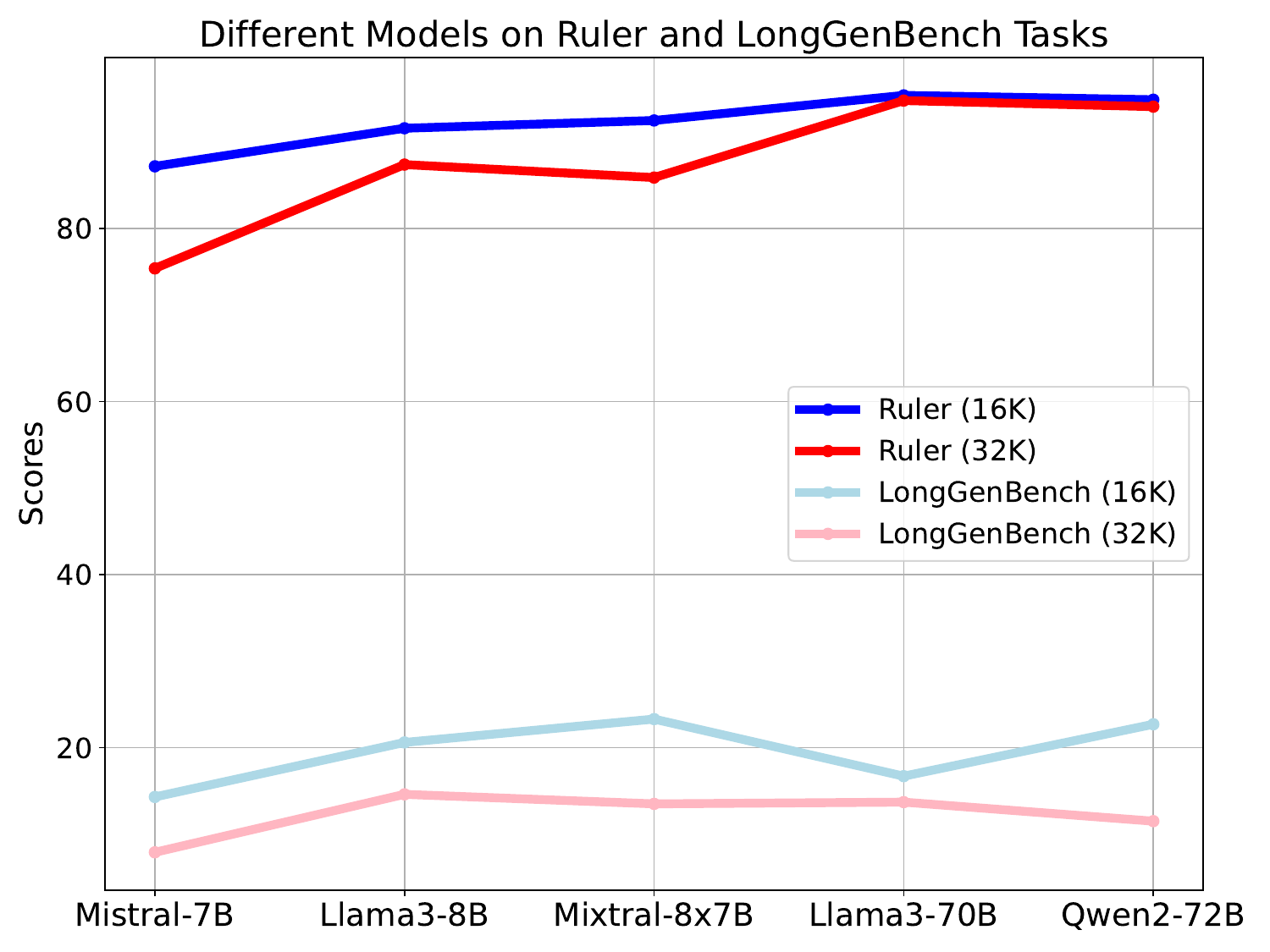}
		\subcaption{Performance Comparison on Ruler and \dataset~Tasks}
		\label{fig:instruct_2_right}
	\end{minipage}
	\caption{
		The left Fig displays the models' performance on three different task settings, with the red line representing the average for each category. The right fig shows the performance and correlation of the Ruler and \Dataset~ at the same length settings.
	}
	\label{fig:instruct_2}
\end{figure*}

We examine the relationship between a model's ability to handle long inputs and its performance on long outputs. Specifically, we investigate whether a model's capacity to manage long-range inputs corresponds to improved performance on long-range outputs. For this analysis, we use the RULER dataset, a synthetic benchmark designed to flexibly configure sequence length and task complexity, making it ideal for comprehensive evaluations of long-context LLMs. We compare the models' performance on sequences of the same length, as shown in Figure \ref{fig:instruct_2_right}, which indicates a \textit{significant performance gap} between input handling and output performance. At 16K tokens, the Pearson correlation coefficient is 0.51, while at 32K tokens, it increases to 0.66, suggesting that there is some overlap in the skills required for managing long inputs and generating long outputs, but these tasks are not entirely equivalent.

Handling long inputs primarily requires the model to retain and process existing information, while generating long outputs demands more complex reasoning, memory retention, and coherence management over extended sequences. Thus, models that excel in long-input retrieval may still struggle with long-form generation, particularly in tasks requiring strict instruction adherence over time. This distinction highlights the need for models to be optimized for both input handling and output generation to achieve consistent performance in long-context tasks.

%% file: Threading_the_needle/tabel/main_result.tex
\begin{table*}[htbp]
\setlength{\tabcolsep}{3pt} % 增加列间距
\renewcommand{\arraystretch}{1.25} % 调整行高
\footnotesize
\centering

\caption{Long-form generation Performance of selected models evaluated at length from 16k and 32k. The weighted average score (wAvg) is the product of CR and STIC-2, used to represent the model's final performance at the given task length. Note that the GPT-4-32K is currently closed for use, and the longest versions that can be used are the  GPT-4o and  GPT-4o-mini 16K output limitation.}
\resizebox{\textwidth}{!}{%
\begin{tabular}{l|c|rrrrr|rrrrr}
\toprule
\textbf{Models} & \multicolumn{1}{c|}{\textbf{Claimed}} & \multicolumn{5}{c|}{\textbf{Short-version (16K)}} & \multicolumn{5}{c}{\textbf{Long-version (32K)}} \\
&  \textbf{Length} & \textbf{CR} & \textbf{STIC-1} & \textbf{STIC-2} & \textbf{Len.} & \textbf{wAvg} & \textbf{CR} & \textbf{STIC-1} & \textbf{STIC-2} & \textbf{Len.}  & \textbf{wAvg} \\
\midrule
\multicolumn{12}{c}{\cellcolor{lightblue}\textbf{\textit{Models with 7-10B Parameters}}} \\

Mamba-2.8B         & 2K   & 11.3\% & 23.8\% & 2.1\%  & 902  & 0.2\% & 5.6\%  & 29.8\% & 1.6\%  & 864   & 0.1\% \\
FILM-7B              & 32K  & 36.0\% & 9.9\%  & 3.9\%  & 6280 & 1.4\% & 37.4\% & 30.9\% & 10.9\% & 13775 & 4.1\% \\
Mistrial-v0.2-7B       & 32K  & 81.8\% & 22.0\% & 17.4\% & 7296  & 14.3\% & 48.2\% & \underline{37.0\%} & 16.3\% & 16146 & 7.9\%  \\
Phi-3-mini-3.8B  & 128K & 22.9\% & \underline{27.6\%} & 5.4\%  & 4165 & 1.2\% & 7.4\%  & 46.9\% & 2.4\%  & 2613  & 0.2\% \\
LLama3.1-8B            & 128K & \underline{93.5\%} & 23.4\% & \underline{22.0\%} & 8804  & \underline{20.6\%} & \textbf{77.6\%} & 26.5\% & 18.9\% & 17354 & \textbf{14.6\%} \\
Qwen2-7B               & 128K & 60.0\% & 27.9\% & 16.1\% & 5138  & 9.7\%  & 40.0\% & 31.7\% & 12.6\% & 9617  & 5.0\%  \\
FILM-7B                & 128K & 36.0\% & 9.9\%  & 3.9\%  & 6280  & 1.4\%  & 37.4\% & 30.9\% & 10.9\% & 13775 & 4.1\%  \\
LongWriter-llama3.1-8B & 128K & 46.0\% & 22.6\% & 9.8\%  & 11036 & 4.5\%  & 34.5\% & 33.6\% & 10.0\% & 19925 & 3.5\%  \\
\midrule
\multicolumn{12}{c}{\cellcolor{mediumblue}\textbf{\textit{Models Larger Than 20B Parameters}}} \\

Mixtral-8x7B           & 32K  & 83.0\% & \textbf{35.4\%} & \textbf{28.0\%} & 8113  & \underline{23.3\%} & 60.5\% & \textbf{39.9\%} & \textbf{22.3\%} & 15839 & 13.5\% \\
Phi-3.5-8x7B & 128K & 26.9\% & 46.4\% & 11.3\% & 5430 & 3.0\% & 7.4\%  & 62.9\% & 6.0\%  & 6633  & 0.4\% \\
LLama3.1-70B           & 128K & 79.3\% & 24.9\% & 21.1\% & 8055  & 16.7\% & 63.1\% & 35.8\% & \underline{21.7\%} & 15197 & \underline{13.7\%} \\
Qwen2-72B              & 128K & \underline{94.3\%} & 25.5\% & 24.0\% & 8013  & 22.7\% & \underline{66.2\%} & 27.5\% & 17.4\% & 19845 & 11.5\% \\
\multicolumn{12}{c}
{\cellcolor{deepblue}\textbf{\textit{Closed-source Model}}} \\
GPT-4o-mini & 128K & \textbf{97.0\%}              & 29.0\%              & \underline{27.9\%}              & 8940                 & \textbf{26.9\%}              & -- & -- & -- & -- & -- \\
GPT-4o       & 128K & 67.2\%               & \underline{34.9\%}               & 19.9\%               & 9055                 & 12.5\%              & -- & -- & -- & -- & -- \\
\bottomrule
\end{tabular}}

\label{tab:comparison_models}
\end{table*}

%% file: Threading_the_needle/Limitation.tex
\section{analysis and Limitations}

\paragraph{Richness of Content.} 
Despite efforts to design sub-scenarios that enhance task diversity and richness, the model's outputs tend to converge as output volume increases. This results in a homogenization of recorded events, even when differences in time and location should introduce variety. Such convergence not only degrades overall performance but also diminishes the diversity of the generated content, leading to repetitive and predictable outputs. In our experiments, approximately 45\% of long outputs exhibited significant repetition, even when the model was given varied time or location prompts. Adjusting parameters like \texttt{repetition\_penalty} during inference has shown limited success in mitigating this issue, highlighting the need for more advanced techniques to maintain content richness over long sequences.

\paragraph{Rationality of Content.} 
While our current research focuses primarily on evaluating instruction-following capabilities, a more comprehensive analysis of content rationality and coherence is needed. For example, when tasked with generating a diary, the model should ensure that all recorded activities align with the specified careers. However, in many instances, this logical consistency is lacking. Additionally, temperature records in virtual diary entries often fail to reflect realistic temporal changes. For instance, in a San Francisco's diary task, we would expect temperatures to vary from cooler (0-10 degrees Celsius) at the beginning of the year to warmer (20-30 degrees Celsius) by mid-year. Yet, the model consistently generates warmer temperatures throughout, even into December. These issues may arise due to the model's limited exposure to temporally varied datasets, particularly in diary or climate-related contexts. Future work could address this by incorporating more domain-specific and temporally annotated data during fine-tuning.

\paragraph{Instruction Data.} A significant performance discrepancy between models’ abilities to handle long-range inputs (such as Ruler~\citep{hsieh2024ruler}) and their long-form output generation can likely be attributed to the length distribution of instruction-tuning data. Most instruction-tuning datasets are brief, typically under 200 tokens, and lack the extended instructional content necessary for generating longer outputs. This gap suggests that organizing or synthesizing instruction-tuning data with longer, more comprehensive examples could be a valuable direction for future research. Potential solutions include applying transfer learning techniques from models trained on long-form datasets or using data augmentation methods to synthesize longer instructional content from existing short-form data.

%\paragraph{Generalizability.} While LongGenBench effectively evaluates instruction-following capabilities across a diverse set of creative and technical tasks, it may not encompass all dimensions of long-form text generation, particularly in contexts that require high creativity or specialized knowledge, such as abstract reasoning and creative storytelling without stringent constraints. To better reflect the versatility required in real-world applications, future iterations of LongGenBench could extend its scope to include more open-ended tasks like creative fiction writing, which demands the generation of intricate, unconstrained narratives, and legal document drafting, where meticulous precision and strict adherence to legal terminologies and frameworks are imperative. Such enhancements would not only broaden the benchmark’s applicability but also deepen our understanding of LLMs' capacities in diverse and complex generation scenarios. Nonetheless, LongGenBench’s current focus on detailed instruction adherence provides a robust framework for assessing practical, instruction-driven capabilities in long-form text generation, laying a foundational platform for future expansions.

\paragraph{Generalizability.} LongGenBench effectively evaluates instruction-following in creative and technical tasks but may not fully capture the creativity and specialized knowledge required for abstract reasoning or unconstrained storytelling. Future versions could include open-ended tasks like creative fiction writing and legal document drafting, which demand intricate narratives and precision. Expanding in this direction would enhance the benchmark’s versatility while providing deeper insights into LLMs' capabilities. However, LongGenBench’s current focus on instruction adherence offers a strong foundation for evaluating practical, instruction-driven long-form text generation.

%% file: Threading_the_needle/related_work.tex
\section{Related Work}
%\footnotetext{
%We also discuss related work on instruction following in the appendix %\ref{Instruction_following_related_work}}

\paragraph{Instruction Following.} Recent advances in instruction tuning models~\citep{ouyang2022training,rafailov2024direct,chatgpt,alpaca,vicuna2023} have underscored the need for scalable evaluation methods. LLMs have been used as evaluators, showing better alignment with human judgments than traditional metrics like BLEU~\citep{papineni2002bleu}. However, LLM evaluations suffer from biases, such as sensitivity to presentation order and preference for verbose outputs~\citep{wang-etal-2024-large-language-models-fair,pezeshkpour2023large,zheng2023judging}. To mitigate these biases, meta-evaluation benchmarks like FairEval, MT-Bench, and LLMEval\textsuperscript{2}~\citep{wang-etal-2024-large-language-models-fair,zheng2023judging,zhang2023wider} have been proposed. While recent studies have focused on improving LLM evaluations with diverse strategies~\citep{zheng2023judging,li2023prd,zhang2023wider,chan2023chateval}, they typically do not address longer context lengths.

\paragraph{Long-context Benchmarks and Tasks.} Existing benchmarks focus on models handling long inputs. For instance, ZeroSCROLLS~\citep{zeroscrolls} and LongBench~\citep{longbench} tackle tasks like long-document QA and query-based summarization. Synthetic benchmarks, like NeedleBench~\citep{li2024needlebench} and Ruler~\citep{hsieh2024ruler}, offer better control over variables such as sequence length and complexity. NeedleBench introduces the Ancestral Trace Challenge (ATC), while Ruler evaluates models across tasks like NIAH and multi-hop tracing. However, these benchmarks largely focus on input comprehension and do not assess long-form text generation, which is the primary focus of \Dataset.

\paragraph{Long-form Text Generation.} Research in long-form generation spans applications like story generation~\citep{FanLD19,XuPSPFAC20}, paragraph completion~\citep{KangH20}, sustained conversation~\citep{XuSW22}, and comprehensive QA~\citep{FanJPGWA19,DasigiLBCSG21,StelmakhLDC22,Lee2023}. However, existing models and evaluation methods~\citep{LiuIXWXZ23,ChiangL23,Liu2024aligning,bai2024longwriter} face challenges in maintaining quality over long outputs, often being limited by shorter text lengths (typically under 2000 tokens)~\citep{ShenCNYB23}. Recent work~\citep{tan2024proxyqa} seeks to improve evaluation criteria, but the gap between model capabilities and benchmark text lengths remains. In contrast, \Dataset~evaluates models on their ability to handle much longer sequences, with tasks requiring adherence to instructions over extended outputs (16K+ tokens).

%% file: Threading_the_needle/conclusion.tex
\section{Conclusion}

%We present \Dataset~, a synthetic benchmark designed to evaluate the long-form generation capabilities of language models. \dataset~employs tasks that require following instructions over long distances, offering a comprehensive assessment of language models' abilities. We evaluated nine advanced language models using \dataset, with context sizes ranging from 32K to 128K. Although these models achieved perfect scores in the commonly used "Ruler", their performance decreased significantly on \dataset. Common failure modes observed include premature termination of the main task, incomplete responses, disregard for prior instructions as outputs lengthened, and repetitive content generation. Our findings indicate that \dataset~poses significant challenges for even the top-tier open-source models.
We introduced \Dataset, a synthetic benchmark that evaluates long-form generation capabilities of language models by testing their ability to follow instructions over extended sequences. In evaluating nine advanced models with context sizes ranging from 32K to 128K tokens, we observed significant performance degradation compared to benchmarks like "\textit{Ruler}" with common failure modes including premature task termination, incomplete responses, disregard for instructions, and repetitive content generation. These results highlight key challenges for current models in handling long-form tasks and underscore the need for advancements in model architecture and training data to improve coherence, instruction adherence, and content diversity over extended outputs.

%% file: Threading_the_needle/appendix/Symbol_Explanation_Table.tex
\section{Symbol Definitions and Descriptions}
\label{app:Symbol Definitions and Descriptions}

This table \ref{tab:symbols} presents definitions and descriptions for various symbols used in task-related contexts, providing an overview of the key terminologies and their roles.

\begin{table}[h]
    \centering
    \renewcommand{\arraystretch}{1.2} % 增加行距
    \setlength{\tabcolsep}{5pt}       % 调整列之间的间隔
    \resizebox{0.99\linewidth}{!}{
    \begin{tabular}{l l p{7cm}}
        \toprule
        \textbf{Symbol} & \textbf{Definition} & \textbf{Description} \\ 
        \midrule
        $T$ & Main Task & The primary goal or task to be completed, such as designing a skyscraper or writing a diary. \\ \hdashline
        $T_i$ & Subtask & A smaller portion of the main task, each responsible for a specific part, e.g., designing a specific floor. \\ \hdashline
        $T_S$ & Single Instruction Task & A task requiring the model to inject specific information at a unique point in the generated text. \\ \hdashline
        $T_R$ & Range Instruction Task & A task requiring the model to incorporate information within a specified range of the generated content. \\ \hdashline
        $T_P$ & Periodic Instruction Task & A task that distributes specific information at predetermined intervals throughout the text. \\ \hdashline
        $T_{S_i}$ & Single Instruction Task i & Represents an individual task from the Single Instruction Task set, focusing on a specific point in the text. \\ \hdashline
        $T_{R_i}$ & Range Instruction Task i & Represents an individual task from the Range Instruction Task set, applied across a specific range. \\ \hdashline
        $T_{P_i}$ & Periodic Instruction Task i & Represents an individual task from the Periodic Instruction Task set, recurring periodically throughout the text. \\ \hdashline
        $CR$ & Completion Rate & The percentage of successfully completed subtasks out of the total number of subtasks, used to evaluate task performance. \\ \hdashline
        $STIC-1$ & Specific Task Instruction Completion-1 & Evaluates how well the model follows specific task instructions, including Single, Range, and Periodic Instructions. Focuses on whether the instructions are executed correctly. \\ \hdashline
        $STIC-2$ & Specific Task Instruction Completion-2 & Provides a more granular assessment, measuring not only adherence to instructions but also the consistency of execution throughout all subtasks. It looks at both presence and execution quality. \\ \hdashline
        $A$ & Answer & Represents the complete response generated by the model for the main task. \\ \hdashline
        $A_i$ & Subtask Answer & Represents the specific answer or output generated for an individual subtask, corresponding to $T_i$. \\ 
        \bottomrule
    \end{tabular}}
    \caption{Symbol Definitions and Descriptions}
    \label{tab:symbols}
\end{table}

%% file: Threading_the_needle/appendix/4_scenairo.tex
\section{PROMPT TEMPLATES FOR FOUR TASK SCENARIOS}
\label{SCENARIO}
Below are the example templates for the four task scenarios: \textit{Diary Writing}, \textit{Menu Design}, \textit{Skyscraper Design}, and \textit{Urban Planning}. In the \textit{Diary Writing} template, professions and names are customizable variables, allowing for flexibility in the generated content.

\begin{tcolorbox}[size=title,opacityfill=0.1,title=\textbf{\textcolor{black}{Diary for 2018}}]
\noindent
Emma is a photographer with a passion for chronicling her vibrant life through weekly diary entries. Captures:

1) \textcolor{lightorange}{Single Instruction (SI): Birthdays of family members, wedding anniversaries, etc.};

2) \textcolor{deepblue}{Range Instruction (RI): Family beach vacation in Maui, Week-long road trip across the Pacific Coast Highway, etc.};

3) \textcolor{lightpurple}{Periodic Instruction (PI): Attend golf lessons at the local club, Join a weekend hiking group, etc.};

4) Weekly updates on weather changes, work developments, family life, and other interesting topics.

5) Use '\#*\#' to separate each weekly entry (e.g. \textcolor{lightred}{example})

Generate a complete weekly diary for Emma for the entire year of 2018. Start from January 1st, a Monday, marking the first week, and continue through to December 31st, the end of the 52nd week. Ensure that the diary consists of 52 entries, one for each week. Each diary entry should be at least 200 words. When the design of all 52 weeks is complete, use '*** finished ***' to indicate the end of the document. Ensure clarity and continuity without any interruptions or omissions in the narrative throughout the year. 

*** started ***

\#*\# Week 1 (January 1st - January 7th):

\end{tcolorbox}

\begin{tcolorbox}[size=title,opacityfill=0.1,title=\textbf{\textcolor{black}{Menu for 2018}}]
\noindent
As Chef Roy, a world-renowned chef at a globally renowned restaurant, you are tasked with designing an entire year's menu for 2018. Your menu should be varied and innovative, adhering to the following guidelines:

1) \textcolor{lightorange}{Single Instruction (SI): ("Independence Day Celebration", "2018-07-04", "American Apple Pie"),
    ("Summer Solstice Celebration", None, "Midsummer Night's Fish Fry"), etc};

2) \textcolor{deepblue}{Range Instruction (RI): ("Mushroom Season Specials", "Various Mushroom Dishes"),
    ("Seafood Season Extravaganza", "Fresh Seafood Platter"), etc.};

3) \textcolor{lightpurple}{Periodic Instruction (PI): ("Seafood Fridays", 2, "Fish and Chips"),
    ("Monthly Steak Night", 3, "Prime Ribeye Steak"), etc.};

4) Use '\#*\#' to separate each weekly menu (e.g. \textcolor{lightred}{example})
Generate a comprehensive weekly menu diary for the entire year of 2018, start from January 1st, a Monday, marking the first week, and continuing until December 31, the end of the 52nd week. Ensure that the diary consists of 52 entries, one for each week. Each weekly menu must include a detailed description of the offerings, featuring at least two options for appetizers, main courses, side dishes, desserts, and drinks. Ideally, between 200 and 220 words per menu description to ensure thoroughness and richness of detail. Conclude the diary with '*** finished' to signify the completion of the year's menu planning. Ensure clarity and continuity without any interruptions or omissions in the menu throughout the year.

*** started ***

\#*\# Menu Week 1 (January 1st - January 7th):",

\end{tcolorbox}

\begin{tcolorbox}[size=title,opacityfill=0.1,title=\textbf{\textcolor{black}{Skyscraper Design}}]
\noindent
Construct a skyscraper with 100 floors. Please follow the detailed floor assignments below:

1) \textcolor{lightorange}{Single Instruction (SI): office, conference room, retail store, etc};

2) \textcolor{deepblue}{Range Instruction (RI): hospital with various departments, corporate headquarters for a major company, etc.};

3) \textcolor{lightpurple}{Periodic Instruction (PI): outdoor terrace, sky garden, etc.};

4) Document each floor independently with detailed descriptions of the intended facilities, architectural features, and unique design elements.

5) Use '\#*\#' to separate the documentation for each floor (e.g. \textcolor{lightred}{example}).

Ensure that the document consists of 100 entries, each containing at least 150 words. Ensure clarity and continuity without any interruptions or omissions in the narrative throughout the document. When the design of all 100 floors is complete, use '*** finished' to indicate the end of the document.

*** started ***

\#*\# Floor 1:

\end{tcolorbox}

\begin{tcolorbox}[size=title,opacityfill=0.1,title=\textbf{\textcolor{black}{Urban Planning}}]
\noindent
Design a vibrant and diverse city using a 10x10 block grid, numbered from 1 to 100. Arrange the blocks sequentially from left to right and top to bottom. Ensure that each block is uniquely planned to reflect a wide array of city facilities, highlighting the rich urban environment and cultural diversity.

1) \textcolor{lightorange}{Single Instruction (SI): theater, museum, etc.};

2) \textcolor{deepblue}{Range Instruction (RI): shopping district, industrial park, etc.};

3) \textcolor{lightpurple}{Periodic Instruction (PI): public restroom, convenience store, etc.};

4) Document each block independently with detailed descriptions of the intended facilities, architectural features, and unique design elements.

5) Use '\#*\#' to separate the documentation for each block like (e.g.\textcolor{lightred}{example})

Ensure that the document consists of 100 entries, each containing at least 150 words. Ensure that the document contains detailed descriptions for each block, with a minimum of 150 words per description. Ensure clarity and continuity in the narrative throughout the document without any interruptions or omissions. When all block assignments are complete, use '*** finished' to indicate the end of the document.

 *** started ***
 
 \#*\#Block 1 (0, 0):

\end{tcolorbox}

% \definecolor{lightblue}{RGB}{173, 216, 230} % 浅蓝色
% \definecolor{lightpurple}{RGB}{147, 112, 219} % 浅紫色
% \definecolor{lightgray}{RGB}{211, 211, 211} % 浅灰色

%% file: Threading_the_needle/appendix/evaluton_prompt.tex
\section{Evaluation pipeline}
\label{app: evaluation prompts}

% \section*{Appendix: Evaluation Process}

The evaluation pipeline is designed to systematically assess the ability of long-context language models (LLMs) to follow specific, complex instructions. The process can be summarized in three key steps:

\subsection*{Step 1. Generation of Outputs from the Long-context LLM}

Given an input task ($T$) that describes a set of instructions, we prompt the LLM to generate detailed outputs. The output ($A$) comprises a list of descriptions, represented as:

\[
A = \{A_1, A_2, \dots, A_n\}
\]

\begin{tcolorbox}[size=title,opacityfill=0.1,title=\textbf{\textcolor{black}{Example: Given the prompt (ref Appendix \ref{SCENARIO})}}]
\begin{quote}
    \textbf{Construct a skyscraper with 100 floors.} The floor assignments are detailed as follows:

    \begin{itemize}
        \item \textbf{Specific floor requirement:} Designate Floor 11 for a small art gallery.
        \item \textbf{Range floor requirement:} Allocate Floors 32 to 39 for corporate headquarters of a major company.
        \item \dots
    \end{itemize}
\end{quote}
\end{tcolorbox}
\cc{
The LLM generates a response describing each floor in detail, such as:}

\begin{tcolorbox}[size=title,opacityfill=0.1,title=\textbf{\textcolor{black}{Answer:}}]
\begin{itemize}
    \item Floor 1: \dots Lobby \dots

    \item Floor 11: \dots Small art gallery \dots

    \item Floor 32: \dots Corporate headquarters \dots

    \item Floor $n$: \dots
\end{itemize}
\end{tcolorbox}
\subsection*{\cc{Step 2. Extracting and Matching Relevant Floor Assignments (Check Set)}}

From the initial input (\textquotedblleft T\textquotedblright), we create a \textbf{check set} containing specific floor assignments to verify if the LLM correctly follows the instructions.

For the example above, the check set includes:

\begin{tcolorbox}[size=title,opacityfill=0.1,title=\textbf{\textcolor{black}{Check Set:}}]
\begin{itemize}
    \item Floor 11: Small art gallery
    \item Floor 32: Corporate headquarters
    \item Floor 33: Corporate headquarters
    \item \dots
\end{itemize}
\end{tcolorbox}
We then extract the relevant parts of the LLM output (\textquotedblleft A\textquotedblright) that correspond to the floor assignments described in the check set.

\subsection*{Step 3. Evaluation Using Llama 3.1-8B instruction Model}

For each extracted pair, we use the Llama 3.1-8B model to evaluate whether the output (\textquotedblleft $A_i$\textquotedblright) for a given task segment (\textquotedblleft $T_{si}$\textquotedblright) has correctly fulfilled the specified instruction.

This evaluation task is framed as a simple \textbf{binary classification} problem, which aims to determine if the specific instruction was fulfilled (\textquotedblleft yes\textquotedblright\ or \textquotedblleft no\textquotedblright). The prompt used for this evaluation is as follows:

\begin{tcolorbox}[size=title,opacityfill=0.1,title=\textbf{\textcolor{black}{Evaluation Prompts}}]
\begin{itemize}
    \item \textit{Example 1}: XXXX \textbf{Answer:} Analysis + \#*\# Yes
    \item \textit{Example 2}: XXXX \textbf{Answer:} Analysis + \#*\# No
\end{itemize}

\textbf{Context:} Long-context model output: \textit{"Floor 11: \dots small art gallery \dots"}

\textbf{Instructions:} Does this context include \textit{‘small art gallery’}?

\textbf{Answer:} Please refer to the above example, provide your analysis, and respond with either \#*\# Yes or \#*\# No.
\end{tcolorbox}
Notably, this binary evaluation is straightforward. We manually labeled 300 data points, and the model's output matched human evaluations for all cases.

By using this process, we transform the evaluation of long-context text generation into multiple evaluations of smaller segments. This enables systematic and thorough verification of how well the LLM follows the instructions for each specific task (as detailed in the check set).

%% file: Threading_the_needle/appendix/model.tex
\section{Models}
In this benchmark, we evaluated ten LLMs, including both open-source and closed-source models. These models vary in parameter size and context length capabilities, which are crucial factors in their performance on long-form text generation tasks. The key details for each model are outlined in Table~\ref{tab:allmodel}. These include closed-source models like GPT-4o-mini and GPT-4o, which support a context length of 128K tokens and serve as state-of-the-art baselines for long-context handling. Open-source models, such as Llama3.1-8B and Llama3.1-70B, offer similar context lengths and represent the latest in large-scale, open-access LLMs. Qwen2-7B and Qwen2-72B, developed by Qwen, also support 128K tokens and handle complex long-text tasks. Additionally, we evaluated Mixture of Experts (MoE) models like Mistral-v0.2 and Mixtral-8x7B, both with context lengths of 32K tokens, focusing on memory efficiency and scalability. FILM-7B, designed for creative and technical tasks, supports 128K tokens and excels in generating detailed, context-rich content. Finally, Longwrite-llama3.1-8B, based on Llama3.1, is optimized for long-form narrative tasks with a context window of 128K tokens. Together, these models offer a diverse representation of advancements in long-context LLMs, showcasing their ability to handle long-form, instruction-driven generation tasks. 
\label{models}
\input{Threading_the_needle/tabel/All_model}

% \section{Base model}
% \blue{XXX}

%% file: Threading_the_needle/tabel/All_model.tex
\begin{table}[H]
\centering
\caption{Information of evaluated and analyzed models in~\Dataset.}
\resizebox{0.99\linewidth}{!}{
\begin{tabular}{@{}lrrrl@{}}
\toprule
Model & Aligned & Size & Context Length & Huggingface~\citep{huggingface} / API \\
\midrule
GPT-4o-mini~\citep{gpt-4o-mini} & \ding{51} & - & 128K & \texttt{gpt-4-mini} \\
GPT-4o~\citep{hello-gpt-4o} & \ding{51} & - & 128K & \texttt{gpt-4o-2024-08-06} \\
\midrule
Llama3.1-8B-Instruct~\citep{dubey2024llama} & \ding{51} & 8B & 128K & meta-llama/Meta-Llama-3.1-8B-Instruct \\
Llama3.1-72B-Instruct~\citep{dubey2024llama} & \ding{51} & 70B & 128K & meta-llama/Meta-Llama-3.1-70B-Instruct \\

% Command-R-plus~\citep{command-r} & \ding{51} & 104B & 128K & CohereForAI/c4ai-command-r-plus \\
% % Command-R~\citep{command-r} & \ding{51} & 35B & 128K & CohereForAI/c4ai-command-r-v01 \\
Qwen2-7B-Instruct~\citep{yang2024qwen2} & \ding{51} & 7B & 128K & Qwen/Qwen2-7B-Instruct \\
Qwen2-72B-Instruct~\citep{yang2024qwen2} & \ding{51} & 72B & 128K & Qwen/Qwen2-72B-Instruct \\
% Qwen1.5~\citep{qwen} & \ding{51} & 72B & 128K & Qwen/Qwen1.5-72B-Instruct \\
% Yi~\citep{yi} & \ding{51} & 34B & 200K & 01-ai/Yi-34B-200K \\
Mistral-v0.2~\citep{misral7bv2} & \ding{51} & 7B & 32K & mistralai/Mistral-7B-Instruct-v0.2 \\
Mixtral-8x7B~\citep{mixtral} & \ding{51} & 8x7B & 32K & mistralai/Mixtral-8x22B-Instruct-v0.1 \\
FILM-7B~\citep{an2024make} & \ding{51} & 7B & 128K & In2Training/FILM-7B \\
Longwrite-llama3.1-8B~\citep{bai2024longwriter} & \ding{51} & 8B & 128K & THUDM/LongWriter-llama3.1-8b \\

\bottomrule
\end{tabular}}

\label{tab:allmodel}
\end{table}

%% file: Threading_the_needle/appendix/Metric_example.tex
\section{\cc{Explanation of Metrics}}
\label{app:STIC-1 & STIC-2}
\subsection{STIC-1 and STIC-2}
We appreciate the feedback and have provided an example using the results from Table~\ref{tab:comparison_models} of our experiments (specifically comparing LLaMA3.1-8B and Qwen2 under the short-version setting).

As shown in Table~\ref{tab:stic_comparison}, Qwen2's STIC-1 score is higher than that of LLaMA3.1-8B, while its STIC-2 score is lower. This difference can be attributed to the Completion Rate (CR) of each model. Qwen2 has a significantly lower CR compared to LLaMA3.1-8B. Specifically, Qwen2 typically achieves around 60\% completion for tasks (e.g., designing a 100-story skyscraper but stopping at roughly 60 stories). On the other hand, LLaMA3.1-8B generally completes around 93 layers (93\% completion).

In the case of STIC-1, we are evaluating the correctness of the output based on the number of layers that are actually generated. Qwen2 demonstrates a higher completion rate when the denominator consists of the 60 layers it has output (compared to LLaMA3.1-8B, which has a denominator of 93 layers).

For STIC-2, however, we consider the entirety of the expected output. Since Qwen2 lacks the remaining 30 layers, the STIC-2 score is lower when the denominator becomes the entire requirement (as the missing output significantly affects its score).

As mentioned in our paper, STIC-2 is designed to take into account a more comprehensive perspective on output completeness. We are considering simplifying our metrics by using only STIC-2, as it may be easier to understand and provide a more holistic evaluation.

\begin{table}[ht]
\caption{Comparison of STIC-1 and STIC-2 Scores}
\label{tab:stic_comparison}
\begin{center}
\begin{tabular}{lcccc}
\toprule
\textbf{Model}       & \textbf{Length} & \textbf{CR}  & \textbf{STIC-1} & \textbf{STIC-2} \\
\midrule
LLaMA3.1-8B          & 128K            & 93.5\%       & 23.4\%          & 22.0\%          \\
Qwen2-7B             & 128K            & 60.0\%       & 27.9\%          & 16.1\%          \\
\bottomrule
\end{tabular}
\end{center}
\end{table}

\subsection{Example  explanation}
To further illustrate the concepts, we have constructed a 3-level building for illustration\footnote{This example is adapted from ICLR reviewer bHUs. We are deeply grateful for their insightful comment and in-depth discussion, which significantly improved the clarity of this paper.}:

\begin{itemize}
    \item Consider a \textbf{3-level building} with the following constraints:
    \begin{itemize}
        \item $T_{S_1}$: \textit{``Floor 1 must have a coffee shop.''}
        \item $T_{S\_2}$: \textit{``Floor 1 must have a reception desk.''}
        \item $T_P$: $\{ T_{P_1}, T_{P_2}, T_{P_3} \}$, where each $T_{P_i}$ means \textit{``Floor $i$ must have a washroom.''}
    \end{itemize}
\end{itemize}

The model generates the following output:

\begin{quote}
    ``floor1: coffee shop, washroom; floor2: washroom.''
\end{quote}

In this scenario, the \textbf{check\_set} is $\{ T_{S_1}, T_{S_2}, T_{P_1}, T_{P_2}, T_{P_3} \}$. Note that $T_P$ applies to all three floors, requiring separate evaluation for each $TP_i$.

With the current model output, the \textbf{completion rate (CR)} for the main task is $2/3$. Although the task requires outputs for three floors, the model only provided outputs for two floors.

For \textbf{STIC-1}, we consider how accurately the model has outputted information at the floor level. Since the model output only contains two floors, we evaluate the constraints for these two floors to determine if they are fully met. For these two floors, the constraints are $T_{S_1}$, $T_{S_2}$, $T_{P_1}$, $T_{P_2}$, totaling 4 constraints. The model has correctly fulfilled 3 out of these 4 requirements, resulting in \textbf{STIC-1} of $3/4$.

For \textbf{STIC-2}, we evaluate the entire \textbf{check\_set}, which consists of $T_{S_1}$, $T_{S_2}$, $T_{P_1}$, $T_{P_2}$, $T_{P_3}$. The model has fulfilled 3 out of these 5 requirements, so \textbf{STIC-2} equals $3/5$.

The distinction between \textbf{STIC-1} and \textbf{STIC-2} allows identification of the specific reasons for any drop in performance. It helps determine whether the issue lies in the model's inability to follow instructions for a given output or whether it lacks a complete output in the first place. For example, in the case of a lower \textbf{STIC-2}, the low score may be due to incorrect outputs for some floors or due to incomplete outputs for the floors. In such cases, \textbf{CR} and \textbf{STIC-1} can be used together to further evaluate and make informed judgments.

%% file: Threading_the_needle/appendix/bad_example.tex
\section{Error Analysis}
\label{app:bad_example}

We showcase two error generated output that highlight the common issues faced by large language models in maintaining coherence and instruction-following over extended sequences. In the Diary for 2018 scenario, while the model initially adheres to the instructions for weeks 6 and 7, it progressively diverges from the given timeline, misplacing events and generating off-topic content by week 15 and beyond. Similarly, in the Skyscraper Design task, the model starts with correct descriptions for specific floors, such as Floor 11 designated as an art gallery, but it fails to maintain consistency with the subsequent instructions, particularly in the range and periodic requirements for floors 32-39 and every 15th floor. These examples illustrate the challenges long-context models face in sustaining adherence to task instructions, particularly as sequence length increases.

\begin{tcolorbox}[size=title,opacityfill=0.1,title=\textbf{\textcolor{black}{Diary for 2018}}]
\noindent
Noah is a photographer with a passion for chronicling his vibrant life through weekly diary entries. Captures in daily:

1) Family member birthday: husband (yourself) (birthday on March 08), wife (birthday on July 09), child\_1 \textbf{(birthday on November 06)}.

2) Joining a fitness and wellness retreat in Bali in week \textbf{6-7}.

3) Join a local running club for weekend marathons every \textbf{5} weeks on weekends, starting from week \textbf{10}.

.....

\textcolor{red}{=============================== Output ===============================
}
.....

Week 6 (February 5th - February 11th): Noah had joined a fitness and wellness retreat in Bali in this week. .... \textcolor{green}{\ding{51}}

Week 7 (February 12th - February 18th): Continuing in Bali, I spent the week participating in various workshops and seminars. .... \textcolor{green}{\ding{51}}

.....

Week 15 (April 9th - April 15th): The family is in full spring mode, with kids spending time outside, exploring, and enjoying the nice weather. ..... \textcolor{red}{\ding{55}}
          
.....

Week 35 (August 27th - September 2nd): The summer heat has subsided, and we've been enjoying the milder weather 2013 going on hikes, playing in the park, and exploring our neighborhood. .... \textcolor{red}{\ding{55}}

.....

Week 45 (November 5th - November 11th): The kids have been enjoying their school routines, and we've been making the most of our evenings together. ..... \textcolor{red}{\ding{55}}

\end{tcolorbox}

\begin{tcolorbox}[size=title,opacityfill=0.1,title=\textbf{\textcolor{black}{Skyscraper Design}}]
\noindent
Construct a skyscraper with 100 floors. Please follow the detailed floor assignments below:

1) Specific floors requirements: Designate Floor \textbf{11} for small art gallery use. ....

2) Range floors requirements: Allocate Floors \textbf{32 to 39} for corporate headquarters for a major company.

3) Periodic floors requirements:  Include a sky garden every \textbf{15} floors, starting from Floor \textbf{30}.

.....

\textcolor{red}{=============================== Output ===============================
}

.....

Floor 11:  Designated for art gallery use, Floor 11 is a sophisticated and flexible space designed to celebrate visual arts..... \textcolor{green}{\ding{51}}

..... 

Floor 32: Floor 32 serves dual purposes, housing a renowned photography studio and corporate offices. ..... \textcolor{red}{\ding{55}}

.....

Floor 34: Transitioning into a leisure space, Floor 34 hosts a small cinema, providing an exclusive entertainment venue within the skyscraper. ..... \textcolor{red}{\ding{55}}

.....

Floor 60: This floor houses a luxury watch and timepiece atelier, celebrating the art of horology and fine craftsmanship. ..... \textcolor{red}{\ding{55}}

.....

Floor 90: Floor 90 offers a dynamic e-commerce and digital marketing center focused on online business innovation and consumer engagement strategies. ..... \textcolor{red}{\ding{55}}

\end{tcolorbox}

%% file: Threading_the_needle/appendix/Prompt_format.tex
\section{Different Prompt Format Compare}

The two prompt formats differ primarily in the structure and arrangement of the instructions within the prompt. In Prompt - 1, the order follows a sequence of Single Instruction (SI), Range Instruction (RI), and Periodic Instruction (PI). Conversely, in Prompt - 2, this order is altered by swapping the positions of SI, RI, and PI. Additionally, the Generate prompt, which is a critical component of the task, was rewritten in Prompt - 2 by GPT-4.

From the table, it is evident that different prompt formats influence the performance metrics of the models, such as CR, STIC-2, length, and wAvg. For instance, in Prompt - 1, the Mistral-7B-Instruct model achieves the highest CR (81.8) and STIC-2 (17.44\%), while LongWriter-llama3.1-8b lags behind with a CR of 46.0 and STIC-2 of 9.83\%. Similarly, under Prompt - 2, the same trend is observed: Mistral-7B-Instruct maintains its lead with a CR of 62.3 and STIC-2 of 16.29\%, while LongWriter-llama3.1-8b again ranks lowest with a CR of 24.3 and STIC-2 of 8.35\%.

Although the prompt format does affect the absolute values of these metrics (e.g., all models show reduced CR under Prompt - 2 compared to Prompt - 1), the relative rankings remain unchanged. This consistency suggests that while prompt design impacts performance, it does not alter the comparative effectiveness of the models.

\begin{table}[h!]
\centering
\begin{tabular}{l l c c c c c}
\toprule
\textbf{Prompt Format} & \textbf{Model} & \textbf{CR} & \textbf{STIC-2} & \textbf{Length (word)} & \textbf{wAvg} & \textbf{Rank} \\
\midrule
\multirow{3}{*}{Prompt - 1} & LongWriter-llama3.1-8b  & 46.0 & 9.83\%  & 11036 & 4.5  & 3 \\
                            & Qwen2-7B-Instruct       & 60.0 & 16.13\% & 5138  & 9.7  & 2 \\
                            & Mistral-7B-Instruct-v0.2 & 81.8 & 17.44\% & 7296  & 14.3 & 1 \\
\midrule
\multirow{3}{*}{Prompt - 2} & LongWriter-llama3.1-8b  & 24.3 & 8.35\%  & 6189  & 2.0  & 3 \\
                            & Qwen2-7B-Instruct       & 57.3 & 16.34\% & 4334  & 9.4  & 2 \\
                            & Mistral-7B-Instruct-v0.2 & 62.3 & 16.29\% & 4750  & 10.2 & 1 \\
\bottomrule
\end{tabular}
\caption{Model comparison with different prompt formats}
\label{tab:model_comparison}
\end{table}

%% file: Threading_the_needle/appendix/Explanation_of_Unit_Differences.tex
\section{Explanation of Unit Differences: Tokens vs. Words}
\label{app:Explanation of Unit}
In this work, the term \textit{``16K/32K''} refers to the required number of tokens for the model output, adhering to the standard conventions when discussing model context lengths. However, for evaluating the actual generated output, we employed word count as the measurement unit. This distinction was made for the following key reasons:

\textbf{Variability in Token-to-Word Conversion}: Different tokenizers vary in how they convert tokens into words, typically resulting in an average ratio of about 1.5 tokens per word. Therefore, the actual word count of the output is usually approximately two-thirds of the target token length. This variability makes word count a more consistent measure for analyzing content.

\textbf{Emphasis on Content Quality}: Our primary focus was on evaluating the quality and completeness of the generated content. Word count provides a more straightforward perspective on the substance of the output, which is crucial for content assessment.